\documentclass{article}

\PassOptionsToPackage{numbers, compress}{natbib}
\usepackage[preprint]{neurips_2022}
\usepackage{listings}
\usepackage{amsmath}
\usepackage{pifont}
\usepackage{xcolor}
\usepackage{colortbl}

\usepackage[utf8]{inputenc} 
\usepackage[T1]{fontenc}    
\usepackage[colorlinks=true,citecolor=citecolor, linkcolor=red, filecolor=magenta, urlcolor=magenta]{hyperref}  
\usepackage{url}            
\usepackage{booktabs}       
\usepackage{amsfonts}       
\usepackage{nicefrac}       
\usepackage{microtype}      
\usepackage{algorithm}
\usepackage{algorithmic}
\usepackage{graphicx}
\usepackage{amsmath}
\usepackage{amssymb}
\usepackage{siunitx}
\usepackage{cleveref}
\usepackage{xspace} %
\usepackage{pifont}
\usepackage{diagbox}
\usepackage{adjustbox}
\usepackage{wrapfig}
\usepackage{soul}
\usepackage{ulem}
\usepackage{enumerate}
\usepackage{ctable}
\usepackage{multirow}
\usepackage{subcaption}

\usepackage{listings}
\usepackage{xcolor}

\definecolor{greencolor}{RGB}{0,128,0}        
\definecolor{bluecolor}{RGB}{0,0,255}         
\definecolor{graycolor}{RGB}{128,128,128}     
\definecolor{normalcolor}{RGB}{0,0,0}         
\definecolor{stringcolor}{RGB}{0,139,139}         

\lstdefinelanguage{TypeScript}{
  keywords={interface, type, extends, implements, class, const, async, await, Promise},
  sensitive=true,
  comment=[l]{//},
  morecomment=[s]{/*}{*/},
  morestring=[b]",
  morestring=[b]'
}

\lstdefinestyle{pythonstyle}{
    language=TypeScript,
    basicstyle=\ttfamily\small,
    backgroundcolor=\color{black!5},
    numbers=left,
    numberstyle=\tiny\color{gray},
    numbersep=5pt,
    frame=single,
    rulecolor=\color{black!40},
    keywordstyle=\color{blue},
    commentstyle=\color{green!60!black},
    stringstyle=\color{purple},
    showstringspaces=false,
    breaklines=true,
    tabsize=4,
    captionpos=b,
    title=\lstname
}

\lstdefinestyle{typescripthighlight}{
    language=TypeScript,
    basicstyle=\ttfamily,
    keywordstyle=\color{greencolor},
    commentstyle=\color{graycolor},
    stringstyle=\color{stringcolor},
    showstringspaces=false,
    morekeywords={interface,type,extends,implements,class,const,async,await,Promise},
    classoffset=1,
    morekeywords={Plugin,IPlugin,Action,Evaluator,Provider,Service},
    keywordstyle=\color{bluecolor},
    classoffset=0,
    breaklines=true,
    breakatwhitespace=true,
    keepspaces=true
}

\definecolor{DeltaColor}{rgb}{0.039,0.73,0.71}
\definecolor{SigmaColor}{rgb}{0.98,0.45,0.0}
\definecolor{AlphaColor}{rgb}{0,0,0.8}
\definecolor{BetaColor}{rgb}{0.8,0,0.8}
\definecolor{GammaColor}{rgb}{0.514,0.34,0.224}
\definecolor{EpsilonColor}{rgb}{0.353,0.725,0.906}
\definecolor{GreenColor}{rgb}{0.137,0.573,0.565}
\definecolor{RedColor}{rgb}{0.949,0.275, 0.224}
\definecolor{citecolor}{HTML}{0071bc}

\newcommand{\projectURL}{\href{https://github.com/elizaOS/eliza}{\tt{elizaOS/eliza}}}

\newcommand{\URL}{\href{https://elizaos.ai/}{\tt{https://elizaos.ai/}}}

\newcommand{\method}{Eliza\xspace}

\newcommand{\xmark}{\textcolor{RedColor}{\ding{55}}\xspace}
\newcommand{\cmark}{\textcolor{GreenColor}{\ding{51}}\xspace}

\newcolumntype{x}[1]{>{\centering\arraybackslash}p{#1pt}}
\newcolumntype{y}[1]{>{\raggedright\arraybackslash}p{#1pt}}
\newcolumntype{z}[1]{>{\raggedleft\arraybackslash}p{#1pt}}
\newlength\savewidth

\newcommand{\colorRef}[1]{\textcolor{RedColor}{#1}}
\crefname{figure}{\colorRef{Fig.}}{\colorRef{Figs.}}
\Crefname{figure}{\colorRef{Figure}}{\colorRef{Figures}}
\crefname{section}{\colorRef{Sec.}}{\colorRef{Secs.}}
\Crefname{section}{\colorRef{Section}}{\colorRef{Sections}}
\Crefname{table}{\colorRef{Table}}{\colorRef{Tables}}
\crefname{table}{\colorRef{Tab.}}{\colorRef{Tabs.}}
\newcommand{\logopic}{%
   \raisebox{-0.5ex}{
      \includegraphics[height=2.5ex]{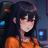}
  }%
}

\newcommand{\logopicagent}{%
   \raisebox{-0.5ex}{
      \includegraphics[height=2.5ex]{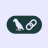}
  }%
}

\newcommand{\logopicautogpt}{%
   \raisebox{-0.5ex}{
      \includegraphics[height=2.5ex]{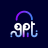}
  }%
}

\newcommand{\logopiccamel}{%
   \raisebox{-0.5ex}{
      \includegraphics[height=2.5ex]{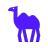}
  }%
}
\title{\logopic \method: A Web3 friendly AI Agent Operating System}

\author{%
Shaw Walters$^{1*}$ \quad Sam Gao$^{1,2*}$ \quad Shakker Nerd$^{1}$ \quad Feng Da$^{1}$ \quad \textbf{Warren Williams}$^{1}$ \\ \quad \textbf{Ting-Chien Meng}$^{1}$ \quad \textbf{Amie Chow}$^{1}$ \quad \textbf{Hunter Han}$^{2}$ \quad \textbf{Frank He}$^{3}$ \quad \textbf{Allen Zhang}$^{4}$ \\ \quad \textbf{Ming Wu}$^{5}$ \quad \textbf{Timothy Shen}$^{6}$ \quad \textbf{Maxwell Hu}$^{7}$ \quad \textbf{Jerry Yan}$^{8}$ \\
$^1$Eliza Labs \quad $^2$AI3 Labs \quad $^3$Heurist AI \quad $^4$GoPlus \quad $^5$Zero Gravity Labs \\ \quad $^6$PipLabs \quad $^7$TownSquareLabs \quad $^8$MIT\\
\texttt{shaw@eliza.systems}
\\ \URL
}

\begin{document}

\maketitle

\def\thefootnote{*}\footnotetext{These authors contributed equally to this work}

\begin{abstract}
    AI Agent, powered by large language models (LLMs) as its cognitive core, is an intelligent agentic system capable of autonomously controlling and determining the execution paths under user’s instructions. With the burst of capabilities of LLMs and various plugins: i.e. RAG, text-to-image/video/3D and etc, the potential of AI Agents has been vastly expanded, with their capabilities growing stronger by the day. However, at the intersection between AI and web3, there is currently no ideal agentic framework that can seamlessly integrate web3 applications into AI agent functionalities. In this paper, we propose Eliza, the first open-source web3-friendly Agentic frameworks that make the deployment of web3 applications effortless. We emphasize that every aspect of Eliza is a regular Typescript program under the full control of its user, and it seamlessly integrates with web3 (i.e. reading and writing blockchain data, interacting with smart contracts and etc). Furthermore, we show how stable performance is achieved through the pragmatic implementation of the key components of Eliza’s runtime. Our code is publicly available at \projectURL.
\end{abstract}

\begin{figure}[bth]
  \centering
   \includegraphics[width=\linewidth]{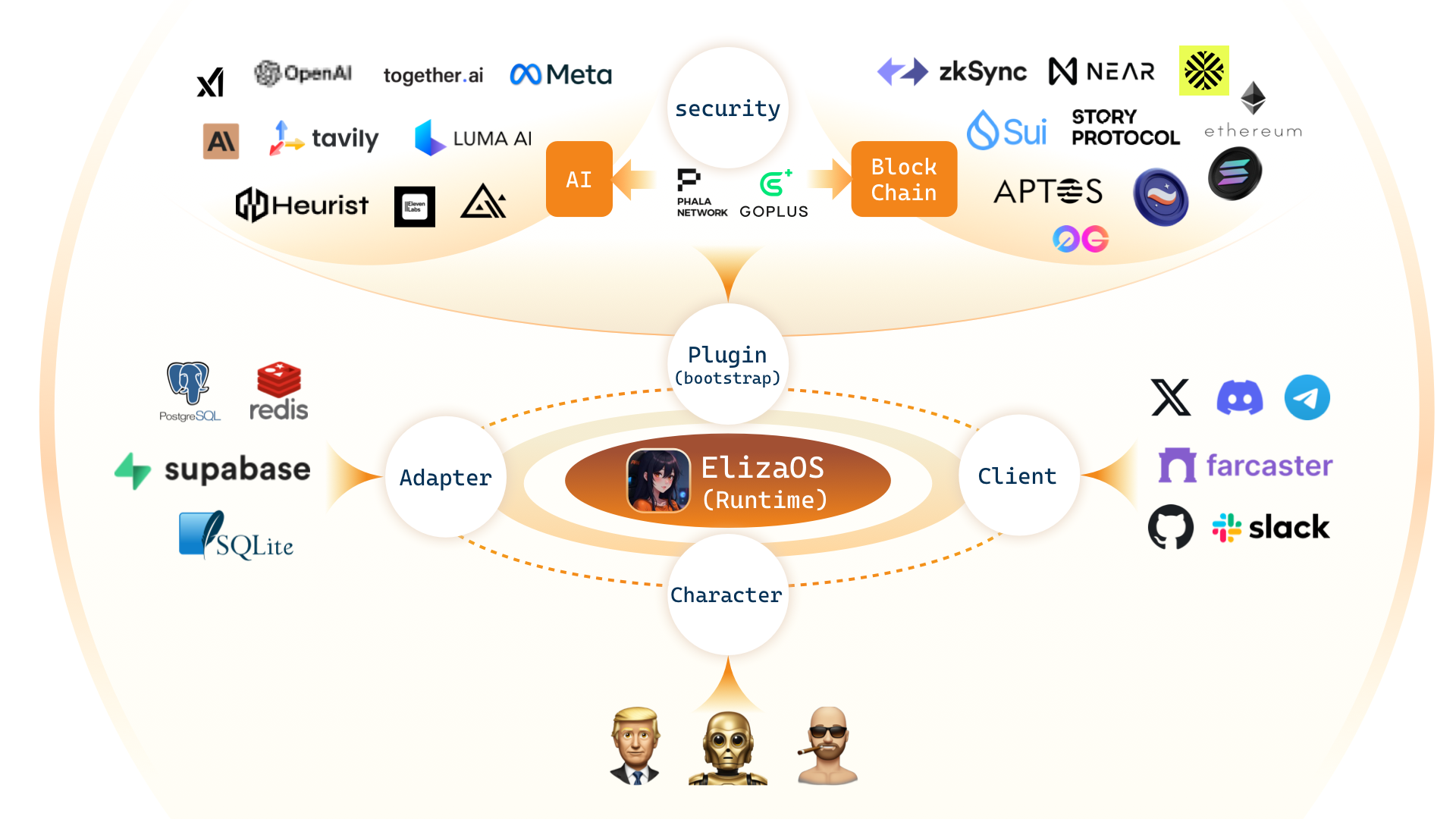}
    \caption{\method is a straightforward yet efficient AI agent operating system, offering a seamless experience for developers to effortlessly launch their first-ever web3-oriented AI Agent.}
    \label{fig:teaser}
\end{figure}

\section{Introduction}
\label{sec: intro}

In the rapidly evolving landscape of AI, the advent of AI Agent, a system driven by large language models (LLMs)~\cite{gpt4,bi2024deepseek,bai2023qwen,touvron2023llama,jiang2023mistral} as its cognitive foundation, marks a significant milestone. This intelligent agentic system is not only capable of autonomously controlling and determining the execution paths under user instructions but also possesses the adaptability to navigate complex tasks with ease. The surge in capabilities of LLMs, coupled with the integration of diverse plugins such as RAG~\cite{graphrag,lewis2020retrieval}, text-to-image/video/3D~\cite{rombach2022high,sora,TripoSR2024} tools, and more, has exponentially expanded the potential of AI Agents (\textit{i.e.} AutoGPT~\cite{yang2023auto}, LangGraph~\cite{LangGraph}, Camel~\cite{li2023camel}, OpenAI Swarm~\cite{openaiswarm} and MiniChain~\cite{rush-2023-minichain}). Their capabilities are advancing at a pace that is nothing short of remarkable, with new functionalities being added and refined on a daily basis.

However, despite the significant advancements in AI technology, a conspicuous gap persists at the confluence of AI and web3. The web3 domain is notably lacking an ideal agentic framework capable of seamlessly integrating web3 applications into its ecosystem, thereby fully unleashing the transformative potential of decentralized AI. This represents a critical void, as the successful integration of AI Agents with web3 technologies has the potential to revolutionize our engagement with decentralized applications and blockchain networks. By doing so, it could pave the way for a more equitable world where the benefits of technological progress are more broadly and fairly distributed among humanity.

In this paper, we introduce Eliza, a pioneering open-source web3-friendly agentic operating system designed to bridge this gap. Eliza is the first of its kind, offering a platform that makes the deployment of web3 applications not only possible but also effortless. We emphasize that every aspect of Eliza is crafted as a regular Typescript program, ensuring that it remains under the full control of its users while also providing seamless integration with web3 functionalities. This includes, but is not limited to, reading and writing blockchain data, interacting with smart contracts, and much more functionality.

Furthermore, we delve into how the key components of Eliza’s runtime are implemented. As shown in \cref{fig:teaser}, we explain how these components are designed to work in harmony, enabling the framework to achieve stable performance while maintaining the flexibility required to adapt to the ever-changing demands of web3 applications. By solving the challenges of integrating AI with web3, Eliza stands at the forefront of a new era in technology, where the possibilities are as boundless as the imagination of its users.




\begin{table}[t]
\caption{Comparison with trending AI Agent frameworks.}
\label{tab:hand_compare}
\centering
\begin{tabular}{l|ccc>{\columncolor{brown!30}}c}
\toprule
 & \logopicagent LangGraph & \logopicautogpt AutoGPT  & \logopiccamel CAMEL & \logopic \method \\
\hline
Multi-Agent System      & \cmark    & \cmark   & \cmark    & \cmark \\
Social Media       & \xmark    & \cmark   & \cmark    & \cmark \\
Web3 Support      & \xmark    & \xmark    & \xmark    & \cmark \\
Human-in-the-Loop & \cmark    & \cmark    & \xmark    & \cmark \\
Github Trending & \xmark    & \cmark    & \xmark    & \cmark \\
\hline
Language  & $\mathtt{Python}$    & $\mathtt{Python}$  & $\mathtt{Python}$    & $\mathtt{TypeScript}$ \\
Workflow      & $\mathtt{Manual}$    & $\mathtt{Manual}$    & $\mathtt{Manual}$    & $\mathtt{Manual,Automatic}$ \\
\bottomrule
\end{tabular}
\end{table}

\section{Background}
\label{sec: back}

\textbf{Decentralized Trading Bots}: At the heart of the crypto or web3 world lies the functionality of trading, such as transferring tokens and participating in Token Generation Events (TGEs), minting NFTs, and swapping tokens through decentralized exchanges (DEXs). With the proliferation of blockchain public chains like $\mathtt{ETH}$, $\mathtt{SOL}$, $\mathtt{BASE}$ and others, managing and operating one's investment portfolio over fragmented blockchains has become increasingly challenging. Individual investors are in dire need of a system to help manage their portfolios and conduct intelligent operations and trades. Platforms like \textbf{GMGN}, \textbf{Dexscreener}, and \textbf{Bull X} have filled this gap to a great extent, but for intermediate to advanced users with customized needs, the basic functionalities of these platforms may fall short.

\textbf{Business Insights}: Secondly, blockchain data itself contains a wealth of crucial information for traders to make decisions. From simple metrics like changes in token holder counts, token prices, market capitalization, and Total Value Locked (TVL), to more advanced indicators such as the proportion of whale accounts, market-maker styles, and candlestick patterns, all can provide effective assistance to different types of cryptocurrency investors. The emergence of AI agents has brought hope for structuring the complex data on blockchains into high-quality insights to aid investors in making wiser decisions. However, extracting data intelligence is a challenging task, and using a general AI Agent framework for this purpose demands a high level of expertise from users. Therefore, there is an urgent need for a Web3-native AI Agent framework to achieve this.

\textbf{Interaction}: Finally, for the Web3 industry, social media platforms like Twitter, Discord, and Farcaster are essential for connecting with users, obtaining cutting-edge information, and making trading decisions. As an increasing number of Key Opinion Leaders (KOLs) flock to these platforms, the information they disseminate becomes more complex and fragmented. Navigating this landscape to acquire organic insights and critically assess the credibility of KOLs is a universal challenge for traders. An exemplary Agent would enable users to sift through the vast information pool, distilling valuable intelligence without succumbing to information overload, and serving as a genuine intermediary in social media interactions with other users or agents.

In consideration of the needs above, Eliza emerges as the premier open-source, web3-friendly AI Agent Operating System, boasting a modular design that empowers developers and users to tailor solutions to their specific requirements. By harnessing the robust capabilities of AI models and a variety of add-ons, Eliza democratizes access to advanced AI functionalities, significantly reducing the barrier to entry for the general public without the need for extensive coding expertise.

\section{Design Principles}
\label{sec: design}

Eliza is a powerful multi-agent simulation framework designed for creating, deploying, and managing autonomous AI agents. It is built using TypeScript and is capable of interacting across multiple platforms. Numerous projects have been developed based on our framework.

Eliza's success is attributed to its integration of the strong demands of web3 into a design that balances utility and ease of use. There are three main principles behind our choices:

\textbf{Put Web3 Developers First} \quad Since web3 primarily utilizes JavaScript/TypeScript, which is the dominant language for web development, Eliza allows developers to easily integrate blockchain functionality into existing web applications and build decentralized applications (dApps) by leveraging familiar tools and frameworks. Eliza should be a first-class member of that ecosystem. It adheres to the commonly established design goals of keeping interfaces simple and consistent, ideally with one idiomatic way of doing things.


\textbf{Pluggable Modular Design} \quad Eliza decouples its structure into a core \textbf{Runtime} along with four key components: \textbf{Adapter} (data), \textbf{Character} (agent personality), \textbf{Client} (message interaction), and \textbf{Plugin} (universal functionality). This design allows developers or users to freely add their own plugins, clients, characters, and adapters as they wish, without worrying about the details within the core Runtime. It makes extension incredibly easy and paves the way for Eliza to support the most model providers (\textit{i.e.} OpenAI, Llama, Qwen and etc.), platform integrations (\textit{i.e.} Twitter, Discord, Telegram and etc.), chain compatibilities (\textit{i.e.} Solana, Ethereum, Ton and etc.), and highly equipped functions (\textit{i.e.} Text2Image/Video/3D, Web Search, TEE and etc.).

\textbf{Roughness is better} \quad Given limited engineering resources and all else being equal, keeping Eliza's internal implementation simple saves time for adding features, adapting to new situations, and keeping pace with advancements in AI and Web3. Therefore, it is better to have a simple but slightly incomplete solution than a comprehensive yet complex and hard-to-maintain design.
\section{Related Works}
\label{sec: related}


As an AI Agent operating system focusing on web3 and social media, we aim to define our position and differentiate ourselves from both industrial AI Agent frameworks (\textit{i.e.} Bedrock (AWS), Swarm (OpenAI), and smolagent (Huggingface) ~\cite{smolagents}]) and academic-oriented projects ~\cite{huang2022language, wang2023voyager, yao2022react, li2023camel}. Specifically, we will mainly discuss plugins and frameworks below.


\subsection{Plugins} Along with the rapid growth of off-the-shelf plugins, the agent's enhancement can be categoried into two principle forms: \textbf{Internal} and \textbf{External}. Internally, the core principle is to tap into the full potential of the LLM itself, yielding more organized and logical answers and alleviating the long-standing issue of hallucination. Representative works within this paradigm include Chain-of-Thoughts (CoT) ~\cite{wei2022chain}, along with its successful descendants: Zero-shot CoT ~\cite{kojima2022large}, Tree-of-Thoughts (ToT)~\cite{yao2024tree}, Graph-of-Thoughts (GoT) ~\cite{besta2024graph}, and Layer-of-Thoughts (LoT) ~\cite{fungwacharakorn2024layer}. CoT introduced step-by-step explanations, ToT allowed branching to explore multiple solutions, and GoT connected reasoning pathways in a network. LoT, released in October 2024, is a hierarchical reasoning AI that organizes thoughts into layers for structured problem-solving. It filters information through layers of constraints to efficiently and transparently find the most relevant solutions.

While "X-of-T" techniques have significantly enhanced the problem-solving prowess of LLMs, paving the way for more intelligent and insightful AI systems, the role of \textbf{external} information is also crucial. Externally, integrating knowledge from various sources greatly enhances an AI agent's ability to solve real-world practical problems. This includes Retrieval Augmented Generations (RAGs)~\cite{lewis2020retrieval, graphrag}, vector databases~\cite{han2023comprehensive}, and web searches~\cite{tavily}. Furthermore, as AI-Generated Content (AIGC) matures, the ability to convert text into images~\cite{rombach2022high, ramesh2021zero}, videos~\cite{yang2024cogvideox,guo2023animatediff}, and 3D models~\cite{poole2022dreamfusion,TripoSR2024} opens up new possibilities for AI agents, adding a fresh dimension to the capabilities of LLMs.

As shown in \cref{fig:teaser}, Eliza offers robust support for a variety of blockchain plugins, encompassing everything from on-chain transactions to Trusted Execution Environments (TEEs). The comprehensive web3 toolkit is designed to be user-friendly and easily extensible, even for junior developers, thus achieving a balance between simplicity and efficiency. Additionally, the integration of social media support broadens the range of application scenarios, which constitutes the primary arena where these web3-oriented agents can actively participate and demonstrate their value.

\subsection{Frameworks}
\label{sec: frame}

\begin{figure}[hpb]
  \centering
  \includegraphics[width=1.0\linewidth]{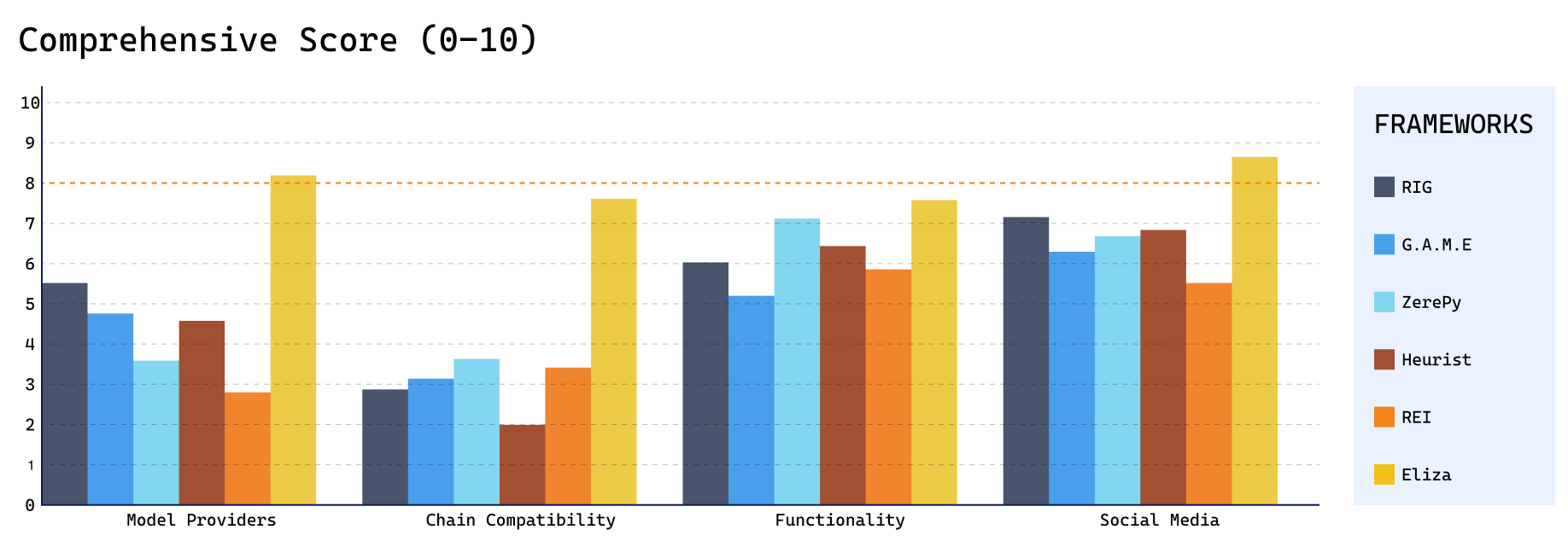}
  \caption{Comparison with AI agent frameworks focuses on web3. Score ranging from 0 (worst) to 10 (best), reflect the views of senior developers come from AI and web3 industry.}
\label{fig:comp}
\end{figure}

AI agent frameworks flourished at the emergence of ChatGPT, has rallied in 2023, where AutoGPT, LangGraph (LangChain) and Camel released their first version on. People from all walks of life have find the potential and benefit in leveraging AI Agent or Workflow to promote their efficiencies in coping with tedious routine jobs.

For web3 industry, due to its highly time-sensitive property and complexity within blockchain interactions, a series of web3-oriented AI Agent frameworks start to emerge: 
\begin{itemize}
    \item \textbf{open source}: RIG, G.A.M.E, ZerePy, Heurist, REI.
    \item \textbf{close source}: Virtual.
\end{itemize}

As an open source framework, Eliza should be compared with its rivals of the same kind: including RIG, G.A.M.E, ZerePy, Heurist, REI. As shown in \cref{fig:comp}, we collect the feedbacks from over 50+ AI researchers and senior blockchain developers to acquire their subjective assessment toward current trending web3 AI agent frameworks, it can be easily observed that Eliza outperforms other frameworks in terms of key indicators: model providers, chain compatibility, functionality and social media.


\section{ElizaOS}
\label{sec: main}



As general frameworks often limited to its highly abstract low-level details and use cases, the direction moving from generic to specialization becomes evident as time goes on. Plus, it is well-known that AI and LLMs are fields that evolve rapidly, with new concepts and ideas emerging every week. Abstractions like LangChain or AutoGPT, which are built around a variety of emerging technologies, find it difficult to withstand the test of time with their framework design.


In the highly time-sensitive web3 industry, developers often need to interact with blockchains for various activities such as transferring tokens, deploying and interacting with smart contracts, and staying updated with the latest information, including cryptocurrency prices, recent statements from Key Opinion Leaders (KoLs), and the holdings of major investors, often referred to as "whales". Almost all of these tasks can be automated through rule-based systems. Prior to the advent of AI Agents, it was challenging to account for all these details and create a comprehensive automated process.

Based on the philosophy derived from previous AI Agent frameworks, we have built a highly controllable and well-orchestrated framework that primarily focuses on the Web3 industry. This framework simplifies the process of bringing powerful AI agents to life by removing hurdles for developers.

\subsection{Core Concepts}

\subsubsection{Agents}

Agents are the core carriers of Eliza that handle autonomous interactions. Each agent runs in a runtime and can interact through various clients (Discord, Twitter, etc.) while maintaining consistent behavior and memory.

From an implementation perspective, \textbf{AgentRuntime} class is the primary implementation of the \textbf{IAgentRuntime} interface, which manages the agent's core functions, including:

\begin{itemize}
    \item \textbf{Message and Memory Processing}: Storing, retrieving, and managing conversation data and contextual memory.
    \item \textbf{State Management}: Composing and updating the agent’s state for a coherent, ongoing interaction.
    \item \textbf{Action Execution}: Handling behaviors such as transcribing media, generating images, and following rooms.
    \item \textbf{Evaluation and Response}: Assessing responses, managing goals, and extracting relevant information.
\end{itemize}

Eliza provides fully functional but not over-designed agent runtime with corresponding state management, memory system and message processing, makes runtime serviced to function in a sound operating state. Here is the minimal code snippet to instantiate a runtime:

\begin{align*}
& \mathtt{const\:\color{blue}runtime \color{black} = new\:\color{red}AgentRuntime\color{black}(\{} \\
& \qquad \mathtt{token: "auth\_token",} \\
& \qquad \mathtt{modelProvider: ModelProviderName.\color{gray}ANTHROPIC \color{black},} \\
& \qquad \mathtt{\color{magenta}character\color{black}: characterConfig,} \\
& \qquad \mathtt{databaseAdapter: new\:DatabaseAdapter(),} \\
& \qquad \mathtt{conversationLength: 32,} \\
& \qquad \mathtt{serverUrl: "http://localhost:7998",} \\
& \qquad \mathtt{\color{magenta}actions\color{black}: customActions,} \\
& \qquad \mathtt{\color{magenta}evaluators\color{black}: customEvaluators,} \\
& \qquad \mathtt{\color{magenta}providers\color{black}: customProviders} \\
& \mathtt{\})}
\end{align*}

\subsubsection{Character Files} Character files are JSON-formatted configurations that define an AI agent's personality, knowledge, and behavior within Eliza. Specifically, Eliza convert a Zod schema (\textbf{CharacterSchema}) into a TypeScript type (\textbf{CharacterConfig}). The basic attributes to define a character are:

\begin{itemize}
    \item Core identity and behavior: Character background, backstory elements and unique character traits.
    \item Model provider configuration: Include but not limited to OpenAI, Anthropic, Llama.
    \item Client settings and capabilities: Blockchain transaction, NFT minting, smart contract deployment.
    \item Interaction and style guidelines: Conversational style, social media post style, knowledge (RAG).
\end{itemize}

By meticulously crafting the character file, users can create an exclusive AI Agent that possesses unique skills and personalities. This process is akin to creating J.A.R.V.I.S. in Iron Man, laying down the most crucial foundation for an autonomous agent.

\subsubsection{Providers}

Providers are essential components that infuse agent interactions with dynamic context and real-time data. Acting as intermediaries, they link the agent to a plethora of external systems, facilitating access to a range of critical information including market data, wallet details, sentiment analysis, and temporal context.

To draw an analogy, providers can be likened to the human perceptual system. Their primary function is to:

\begin{itemize}
    \item Obtain dynamic contextual information
    \item Integrate with the agent runtime
    \item Format information for conversation templates
    \item Maintain consistent data access
\end{itemize}

In Eliza, we have three basic built-in providers: \textbf{Time Provider} (provide temporal context for agent interactions), \textbf{Facts Provider} (maintain conversation facts) and a degen \textbf{Boredom Provider} (manage conversation dynamics and engagement by calculating the boredom level of an agent based on recent messages). Moreover, with built-in registration system, we can mount provider with an instantiated runtime with only one line of code: $$\mathtt{\color{blue}runtime \color{black}.registerContextProvider(customProvider)};$$

\subsubsection{Actions}

Actions serve as the foundational elements within Eliza, dictating the agents' responses and interactions with messages. They empower agents to engage with external systems, adjust their behavior, and execute complex tasks that extend beyond straightforward message exchanges.

An Action encompasses a wealth of functionalities, including but not limited to:

\begin{itemize}
    \item Placing Buy \& Sell Orders
    \item Analyzing PDF documents
    \item Transcribing audio files
    \item Generating NFTs (Non-Fungible Tokens)
\end{itemize}

It's crucial to recognize that the execution of Actions is often pivotal, with financial implications at stake. Each Action must be meticulously designed with a clear and defined purpose. To safeguard against any potential issues, incorporating robust validation mechanisms and comprehensive error handling is not just advisable but essential. These measures are indispensable when configuring user-defined Actions, ensuring the integrity and reliability of the agent's operations in the web3 domain, can be registered through:$$\mathtt{\color{blue}runtime \color{black}.registerAction(customAction)};$$

\subsubsection{Evaluators}

Evaluators represent the final core component of Eliza, tasked with assessing and extracting valuable information from conversations and integrating seamlessly into the \textbf{AgentRuntime}'s evaluation system. Much like Providers, the integration of Evaluators is streamlined and can be executed with a single line of code:$$\mathtt{\color{blue}runtime \color{black}.registerEvaluator(customEvaluator)};$$

In practice, Evaluators empower agents with the ability to:

\begin{itemize}
    \item Build long-term memory
    \item Track goal progress
    \item Extract facts and insights
    \item Maintain contextual awareness
\end{itemize}

In service, Evaluators are dispensible and commonest find under such scenarios: fact extraction to identify key information, goal tracking to monitor progress, and verifying agent functionality under edge cases.

\subsection{Intent Recognition}

\begin{figure}[t]
  \centering
   \includegraphics[width=\linewidth]{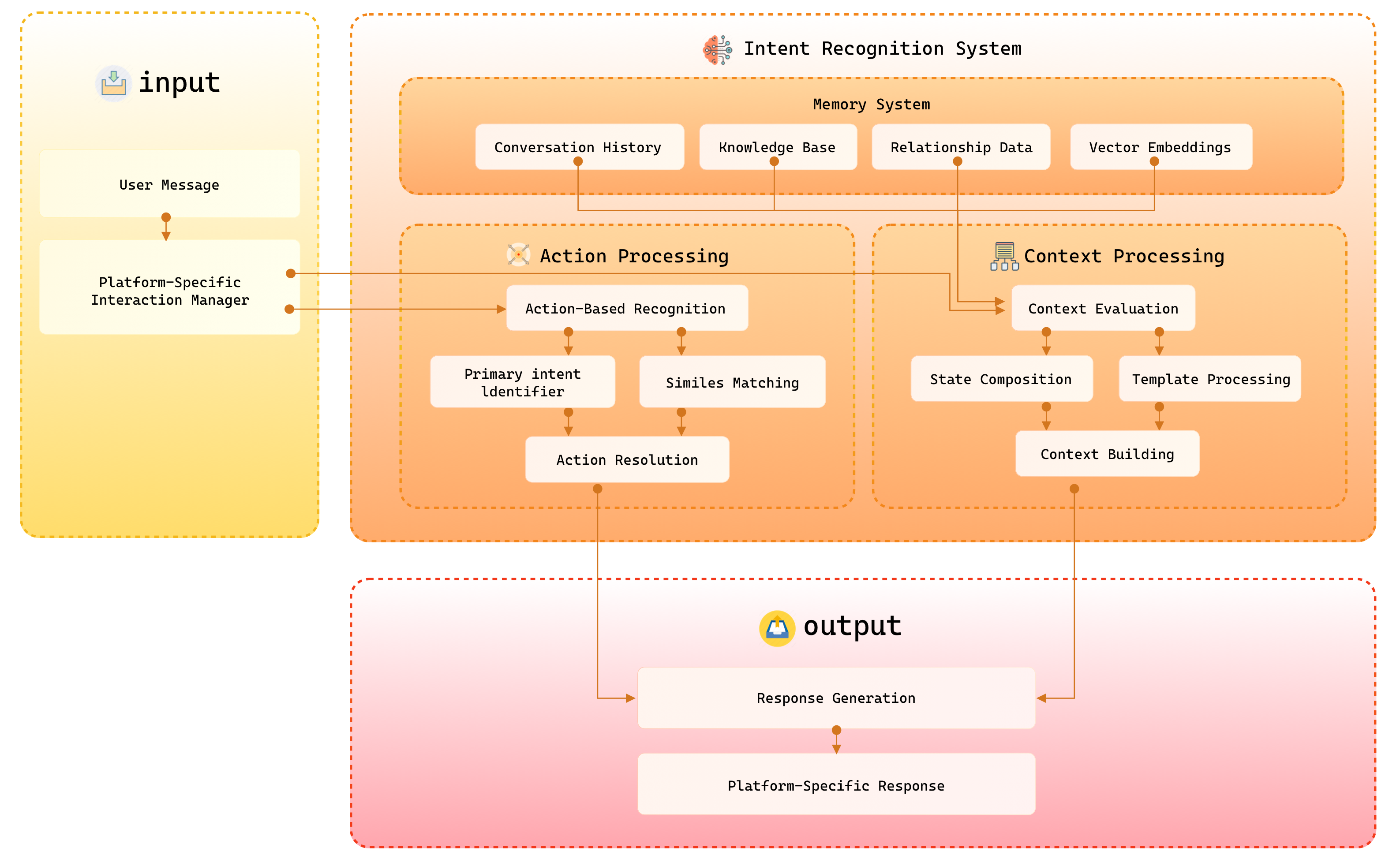}
    \caption{Intent recognition system of Eliza.}
    \label{fig:intent}
\end{figure}

Intent Recognition is the ability of an AI assistant or an AI system, to understand the purpose or goal of a user's request. As shown in \cref{fig:intent}, Eliza employs a multi-layered approach to intent recognition, combining symbolic action definitions with contextual understanding and memory-augmented processing. At its core, the system utilizes a hierarchical action structure, where each intent is defined by a primary identifier accompanied by a collection of semantic similes. This allows for flexible recognition of user intentions across varying linguistic expressions. The primary mechanism is further enhanced by a context-aware evaluation system, which leverages both the immediate conversational state and long-term memory through vector-based retrieval mechanisms.

The framework's intent processing pipeline integrates template-driven context building with platform-specific interaction managers. This enables consistent intent recognition across diverse communication channels while maintaining platform-specific optimizations. The architecture is further augmented by a sophisticated memory system that maintains conversational history, knowledge bases, and relationship tracking. This allows the system to perform contextually relevant intent recognition, adapting to both the immediate conversation flow and the broader interaction history.

The combination of these components results in a robust intent recognition system that can effectively process and respond to user intentions while maintaining contextual awareness and conversational coherence.

\subsection{Plugins}
The Eliza framework implements a flexible plugin architecture that enables modular extension of AI agent capabilities while maintaining system stability and coherence. The plugin system provides a well-defined interface for extending agent functionality through various components:

\paragraph{\ding{182} Media Generation Plugins}
Plugins that enable AI-driven content creation:
\begin{itemize}
\item Image/Video/3D Generation Plugin
    \begin{itemize}
    \item Creates images/videos/3D based on input prompts.
    \item Supports multiple providers (Anthropic, Together and etc).
    \item Image captioning.
    \end{itemize}
\item NFT Generation Plugin
    \begin{itemize}
    \item Generates NFT collections.
    \item Supports various collection attributes.
    \item Integrates with blockchain deployment.
    \end{itemize}
\end{itemize}

\paragraph{\ding{183} Web3 Integration Plugins}
Extensive blockchain support through specialized plugins:

\begin{itemize}
\item Coinbase Plugin Suite:
    \begin{itemize}
    \item Advanced Trading - Complex trading strategies
    \item Commerce Integration - Payment processing
    \item Mass Payments - Bulk transaction handling
    \item Token Contract Management - ERC20/ERC721 deployment
    \item Webhook Integration - Event handling
    \end{itemize}

\item Multi-Chain Support:
    \begin{itemize}
    \item EVM compatibility for Ethereum ecosystem
    \item Solana with trust scoring and wallet management
    \item Additional chains: Aptos, Conflux, Flow, MultiversX, Near, Sui, TON, ICP, zkSync Era
    \item GOAT (Great Onchain Agent Toolkit) integration for cross-chain operations
    \end{itemize}
\end{itemize}

\paragraph{\ding{184} Core Infrastructure Plugins}
Essential services and capabilities:
\begin{itemize}
\item Node Plugin Services:
    \begin{itemize}
    \item BrowserService - Web browsing capabilities
    \item ImageDescriptionService - Image analysis
    \item LlamaService - LLM integration
    \item PdfService - Document processing
    \item SpeechService - Text-to-speech
    \item TranscriptionService - Speech-to-text
    \item VideoService - Video processing
    \end{itemize}
\item TEE (Trusted Execution Environment) Plugin for secure operations
\end{itemize}

The plugin architecture enables:
\begin{itemize}
\item Independent Development - Through clear component interfaces
\item Maintenance Simplification - Through modular package organization
\item Core System Stability - By isolating extensions from core functionality
\item Community Engagement - Through npm package distribution
\item Knowledge Sharing - Via documented examples and TypeScript types
\end{itemize}

This modular architecture allows Eliza to be extended with new capabilities while maintaining a consistent interface and reliable operation across all components. The plugin system's flexibility is particularly evident in its support for both basic utilities like image generation and complex blockchain integrations, making it suitable for a wide range of applications from content creation to decentralized finance.

\subsubsection{Core Architecture}

The plugin components interact through clearly defined interfaces:

\begin{lstlisting}[style=pythonstyle,numbers=none]
/* Core plugin interface for the Eliza framework */
interface Plugin {
    name: string;              // Plugin name
    description: string;       // Plugin description
    actions?: Action[];        // Optional actions
    providers?: Provider[];    // Optional providers
    evaluators?: Evaluator[]; // Optional evaluators
    services?: Service[];     // Optional services
    clients?: Client[];       // Optional clients
}
\end{lstlisting}

\subsubsection{Implementation Patterns}

A minimal plugin implementation follows this pattern:

\begin{lstlisting}[style=pythonstyle,numbers=none]
import { Plugin } from "@elizaos/core";
import { createResourceAction } from "./actions/sampleAction";
import { sampleProvider } from "./providers/sampleProvider";
import { sampleEvaluator } from "./evaluators/sampleEvaluator";

export const samplePlugin: Plugin = {
    name: "sample",
    description: "Enables creation and management of generic resources",
    actions: [createResourceAction],
    providers: [sampleProvider],
    evaluators: [sampleEvaluator],
    services: [],
    clients: []
};
\end{lstlisting}

\subsubsection{Key Benefits}

The plugin architecture offers several key advantages:

\paragraph{1. Modularity}
Plugins encapsulate related functionality through a simple, clear interface as shown in the code snippet above.

\paragraph{2. Extensibility}
New capabilities can be added through standardized component types:

\begin{lstlisting}[style=pythonstyle,numbers=none]
// Example from plugin-image-generation
export const imageGenerationPlugin: Plugin = {
    name: "imageGeneration",
    description: "Generate images",
    actions: [imageGeneration],
    evaluators: [],
    providers: []
};

// Example from plugin-node
export function createNodePlugin() {
    return {
        name: "default",
        description: "Default plugin with basic services",
        services: [
            new BrowserService(),
            new ImageDescriptionService(),
            new LlamaService(),
            new PdfService(),
            new SpeechService(),
            new TranscriptionService(),
            new VideoService(),
            new AwsS3Service()
        ]
    } as const satisfies Plugin;
}
\end{lstlisting}

\paragraph{3. Community Development}
The architecture promotes collaborative development through user-friendly package management, enabling:

\begin{itemize}
\item \textbf{Independent Development:} Well-defined component interfaces (actions, providers, evaluators, services)
\item \textbf{Simplified Maintenance:} Modular package organization
\item \textbf{Core System Stability:} Through isolation of extensions from core functionality
\item \textbf{Community Engagement:} NPM package distribution
\item \textbf{Knowledge Sharing:} Documented examples and TypeScript type definitions
\end{itemize}


\section{Benchmarks}

\subsection{General AI Agent Benchmark}

GAIA~\cite{mialon2023gaia} is a benchmark specifically designed to evaluate the general capabilities of AI agents in solving real-world problems. Successfully answering GAIA questions requires multiple skills, including logical reasoning, multi-modal processing, web browsing, and tool utilization. While these questions are conceptually straightforward for humans, they present significant challenges for current AI systems. We evaluate Eliza's general applicability using this benchmark. In our implementation, we construct swarms of multiple homogeneous agents and employ self-consistency (a prompt-based majority voting mechanism) for final decision-making~\cite{zhugegptswarm}.

\cref{tab:table2} presents Eliza's performance using three agents with a voting system for final decisions. We compare our results against several baselines, including the GPT-series with plugins, GPTSwarm, and other top-ranked methods. The results indicate that Eliza achieves moderate performance compared to these benchmark methods.


\begin{table}
\centering
\caption{Performance on the GAIA Benchmark~\cite{mialon2023gaia}. Eliza demonstrates its generalizability by successfully tackling general-purpose tasks across various levels of difficulty, denote as \%.}
\begin{tabular}{lcccc}
\hline
Method & Level 1 & Level 2 & Level 3 & Average score \\
\hline
GPT-4~\cite{gpt4} & 9.68 & 1.89 & 0 & 4 \\
AutoGPT~\cite{yang2023auto} & 15.05 & 0.63 & 0 & 5 \\
GPTSwarm~\cite{zhugegptswarm} & 30.56 & 20.93 & 3.85 & 18.45 \\
GPT-4o~\cite{gpt4} & 39.78 & 27.04 & 14.58 & 29 \\
Langfun~\cite{langfun} & 58.06 & 51.57 & 25 & 49.33 \\
\hline
\rowcolor{brown!30}
Eliza & 32.21 & 21.70 & 4.36 & 19.42 \\
\hline
\label{tab:table2}
\end{tabular}
\end{table}

\subsection{Web3 Benchmark}

\begin{figure}[htbp]
    \centering
    \includegraphics[width=1.0\linewidth]{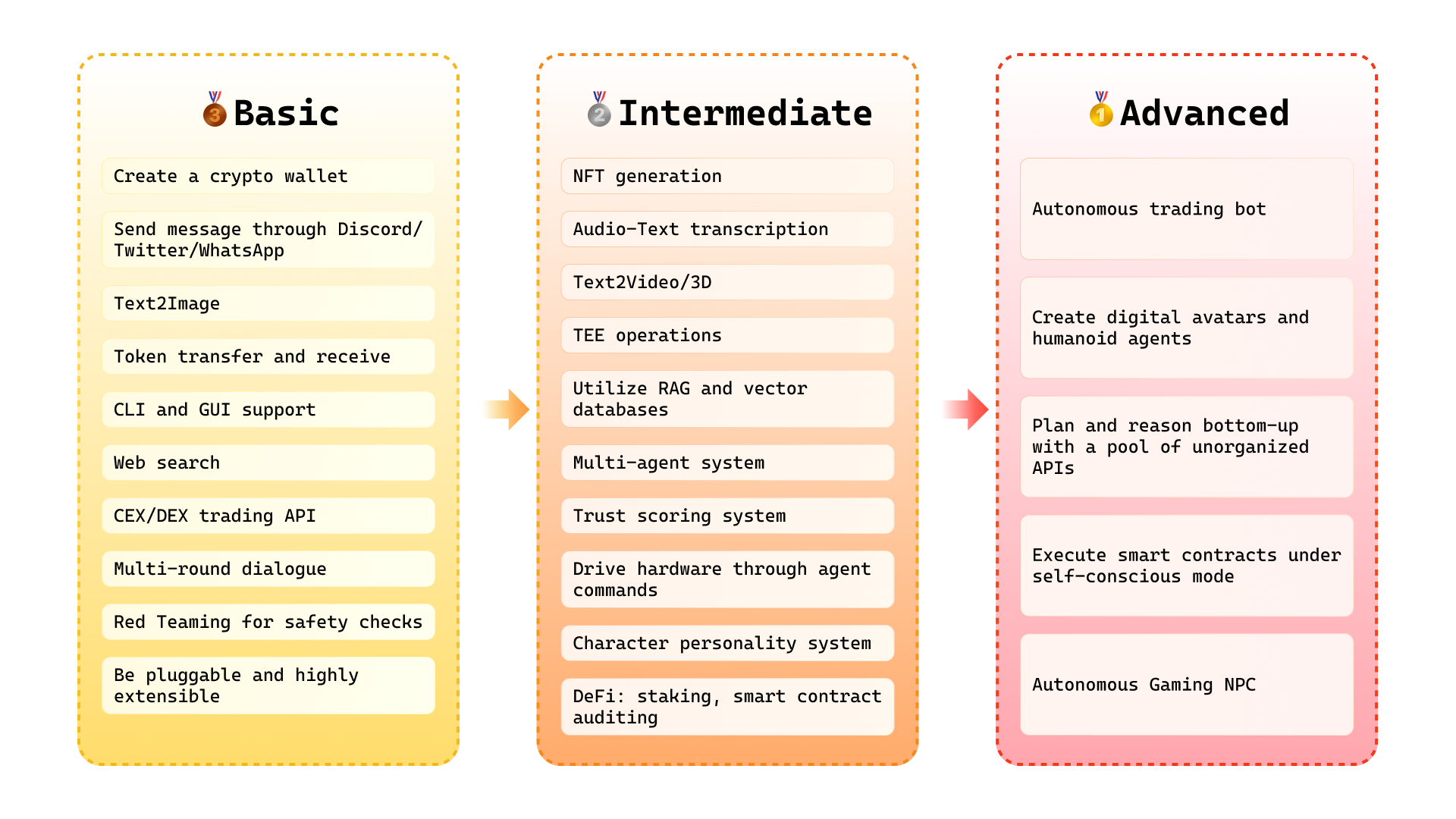}
    \caption{Web3 AI Agent \textbf{'Turing Test'}: An AI agent operating within the Web3 ecosystem is considered to have passed the \textbf{'Turing Test'} if it can successfully manage all tasks categorized as \textbf{Basic}, \textbf{Intermediate}, and \textbf{Advanced} as listed above.}
    \label{fig:benchmark}
\end{figure}

Given that current web3-oriented AI systems are not yet perfected, quantitative results through transplanting Eliza's features on those frameworks are time-consuming and complicated. We hereby establish a foundational standard for existing and forthcoming frameworks, outlining what an AI agent with comprehensive capabilities should encompass. As detailed in \cref{fig:benchmark}, we outline the requirements for testing a web3 AI agent.

For a web3 AI agent, the fundamental requirements must encompass core web3 operations such as creating wallets, transferring and receiving tokens, interacting with smart contracts, and engaging with mainstream social media platforms. Additionally, the agent should support basic trading APIs. Paramount to these functionalities is the thorough red teaming conducted by the AI agent's developers. This exercise is crucial for ensuring the agent's safety and preventing any hazardous behavior that could potentially cause damage to property.

The next phase for AI Agents involves rapidly integrating the latest advancements in AI into the agentic framework. This includes leveraging cutting-edge technologies such as text-to-video/3D, RAG (Retrieval-Augmented Generation) support, and audio-to-text transcription. It is also crucial that this stage addresses web3-validated privacy and safety-related plugins and practices effectively.

The ultimate goal for a fully realized and advanced AI agent is to be capable of autonomous planning and reasoning based on user instructions. Such an agent should be able to automatically devise a suitable execution pipeline without human intervention, drawing from a pool of unorganized APIs. This capability is the key metric for evaluating the intelligence of an AI Agent, mark the agent is already reached its climax.

At present, Eliza is in the midst of transitioning from the primary to the intermediate phase of development. Our team is steadfast in our conviction that we are progressing towards the realization of AI models capable of autonomous action within both digital and physical environments, with the ability to execute complex plans across long-term horizons. We project that this capability will be achieved within the forthcoming years.

Upon the successful operationalization of these "agentic" AI systems, we intend to instantiate multiple units, facilitating their collaborative efforts. This approach aligns with the concept articulated by Dario Amodei, CEO of Anthropic, who envisions a "datacenter of geniuses." The implementation of such a system would represent a paradigm shift in AI capabilities, marking a significant leap forward in the field.

\section{Use Cases}

\subsection{Solana Plugin Example}

The Solana plugin provides functionality for interacting with the Solana blockchain, including token management, swapping, and trust score evaluation, serve as a good example for developer to onboard their blockchain interface with Eliza. 

\subsubsection{Core Features}

\begin{itemize}
    \item Token Management (TokenProvider)
    \item Wallet Integration (WalletProvider)
    \item Trust Score Evaluation (TrustScoreManager)
    \item Token Swapping
    \item FOMO and PumpFun Integration
\end{itemize}

\subsubsection{Key Components}

\paragraph{Token Provider}
Handles token-related operations:

\begin{lstlisting}[style=pythonstyle,numbers=none]
export class TokenProvider {
    async calculateBuyAmounts(): Promise<CalculatedBuyAmounts> {
        const dexScreenerData = await this.fetchDexScreenerData();
        const prices = await this.fetchPrices();
        const solPrice = toBN(prices.solana.usd);
        // ... calculates buy amounts for different conviction levels
        return {
            none: 0,
            low: lowBuyAmountSOL,
            medium: mediumBuyAmountSOL,
            high: highBuyAmountSOL,
        };
    }
}
\end{lstlisting}

\paragraph{Wallet Provider}
Manages wallet interactions:

\begin{lstlisting}[style=pythonstyle,numbers=none]
export class WalletProvider {
    async fetchPortfolioValue(runtime): Promise<WalletPortfolio> {
        const portfolio = await this.getPortfolio(runtime);
        return {
            totalUsd: portfolio.totalUsd,
            totalSol: portfolio.totalSol,
            items: portfolio.items
        };
    }
}
\end{lstlisting}

\paragraph{Trust Score Manager}
Evaluates token and recommender trust scores:

\begin{lstlisting}[style=pythonstyle,numbers=none]
export class TrustScoreManager {
    calculateTrustScore(
        tokenPerformance: TokenPerformance,
        recommenderMetrics: RecommenderMetrics
    ): number {
        const riskScore = this.calculateRiskScore(tokenPerformance);
        const consistencyScore = this.calculateConsistencyScore(
            tokenPerformance,
            recommenderMetrics
        );
        // ... calculates final trust score
        return trustScore;
    }
}
\end{lstlisting}

\subsubsection{Actions}

The plugin provides several key actions:

\begin{lstlisting}[style=pythonstyle,numbers=none]
export const solanaPlugin: Plugin = {
    name: "solana",
    description: "Solana Plugin for Eliza",
    actions: [
        executeSwap,
        pumpfun,
        fomo,
        transferToken,
        executeSwapForDAO,
        take_order,
    ],
    evaluators: [trustEvaluator],
    providers: [walletProvider, trustScoreProvider],
};
\end{lstlisting}

\paragraph{Example Action}
Token Swap (where \textbf{similes} denote alternative names, variations, or synonyms that can trigger the defined Action following intent recognition):
\begin{lstlisting}[style=pythonstyle,numbers=none]
export const executeSwap: Action = {
    name: "EXECUTE_SWAP",
    similes: ["SWAP_TOKENS", "TOKEN_SWAP", "TRADE_TOKENS"],
    handler: async (
        runtime: IAgentRuntime,
        message: Memory,
        state: State,
        _options: { [key: string]: unknown },
        callback?: HandlerCallback
    ): Promise<boolean> => {
        const trustScore = await runtime
            .getProvider('trustScore')
            .evaluateSwap(params);
            
        if (trustScore < runtime.getMinimumTrustThreshold()) {
            return false;
        }
        return true;
    }
};
\end{lstlisting}

\subsubsection{Configuration}

The plugin requires specific environment configuration:

\begin{lstlisting}[style=pythonstyle,numbers=none]
export const solanaEnvSchema = z.object({
    WALLET_SECRET_SALT: z.string().optional(),
    SOL_ADDRESS: z.string().min(1, "SOL address is required"),
    SLIPPAGE: z.string().min(1, "Slippage is required"),
    RPC_URL: z.string().min(1, "RPC URL is required"),
    HELIUS_API_KEY: z.string().min(1, "Helius API key is required"),
    BIRDEYE_API_KEY: z.string().min(1, "Birdeye API key is required"),
});
\end{lstlisting}

\subsection{Advanced Implementation Example: Image Generation}

\paragraph{Core Action Definition}
The plugin defines an image generation action with extensive recognition patterns:

\begin{lstlisting}[style=pythonstyle,numbers=none]
const imageGeneration: Action = {
    name: "GENERATE_IMAGE",
    similes: [
        "IMAGE_GENERATION",
        ...
        "MAKE_A"
    ],
    description: "Generate an image to go along with the message.",
    validate: async (runtime: IAgentRuntime, _message: Memory) => {
        await validateImageGenConfig(runtime);
        
        const anthropicApiKeyOk = !!runtime.getSetting("ANTHROPIC_API_KEY");
        ...
        const falApiKeyOk = !!runtime.getSetting("FAL_API_KEY");
        
        return (
            anthropicApiKeyOk ||
            ...
            falApiKeyOk ||
        );
    }
};
\end{lstlisting}

\paragraph{Provider Validation}
The system implements robust provider validation:

\begin{lstlisting}[style=pythonstyle,numbers=none]
interface IAgentRuntime {
    getSetting(key: string): string | undefined;
}

const validate = async (runtime: IAgentRuntime): Promise<boolean> => {
    const requiredKeys = [
        "ANTHROPIC_API_KEY",
        ...
        "FAL_API_KEY",
    ];

    return requiredKeys.some(key => !!runtime.getSetting(key));
};
\end{lstlisting}

\paragraph{File Management System}
The plugin includes robust file handling capabilities:

\begin{lstlisting}[style=pythonstyle,numbers=none]
import { join } from 'path';
import { existsSync, mkdirSync, writeFileSync } from 'fs';

export function saveBase64Image(
    base64Data: string, 
    filename: string
): string {
    const imageDir = join(process.cwd(), "generatedImages");
    
    if (!existsSync(imageDir)) {
        mkdirSync(imageDir, { recursive: true });
    }
    
    const base64Image = base64Data.replace(
        /^data:image\/\w+;base64,/, 
        ""
    );
    
    const imageBuffer = Buffer.from(base64Image, "base64");
    const filepath = join(imageDir, `${filename}.png`);
    
    writeFileSync(filepath, imageBuffer);
    return filepath;
}
\end{lstlisting}

\paragraph{Image Generation Options}
The system supports comprehensive image generation configuration:

\begin{lstlisting}[style=pythonstyle,numbers=none]
interface ImageOptions {
    width?: number;
    height?: number;
    count?: number;
    negativePrompt?: string;
    numIterations?: number;
    guidanceScale?: number;
    seed?: number;
    modelId?: string;
    stylePreset?: string;
    hideWatermark?: boolean;
}
\end{lstlisting}

\paragraph{Response Handling}
The plugin implements a callback system for handling generated images:

\begin{lstlisting}[style=pythonstyle,numbers=none]
interface Attachment {
    id: string;
    url: string;
    title: string;
    source: string;
    description: string;
    contentType: string;
}

interface Response {
    text: string;
    attachments: Attachment[];
}

interface FileAttachment {
    attachment: string;
    name: string;
}

function handleImageResponse(
    filepath: string, 
    filename: string
): [Response, FileAttachment[]] {
    return [
        {
            text: "Image generated successfully",
            attachments: [{
                id: crypto.randomUUID(),
                url: filepath,
                title: "Generated image",
                source: "imageGeneration",
                description: "AI-generated image",
                contentType: "image/png",
            }],
        },
        [{
            attachment: filepath,
            name: `${filename}.png`,
        }]
    ];
}
\end{lstlisting}



\section{Massive Adoption}
\label{sec: adopt}

\begin{figure}[t]
  \centering
  \includegraphics[width=1.0\linewidth]{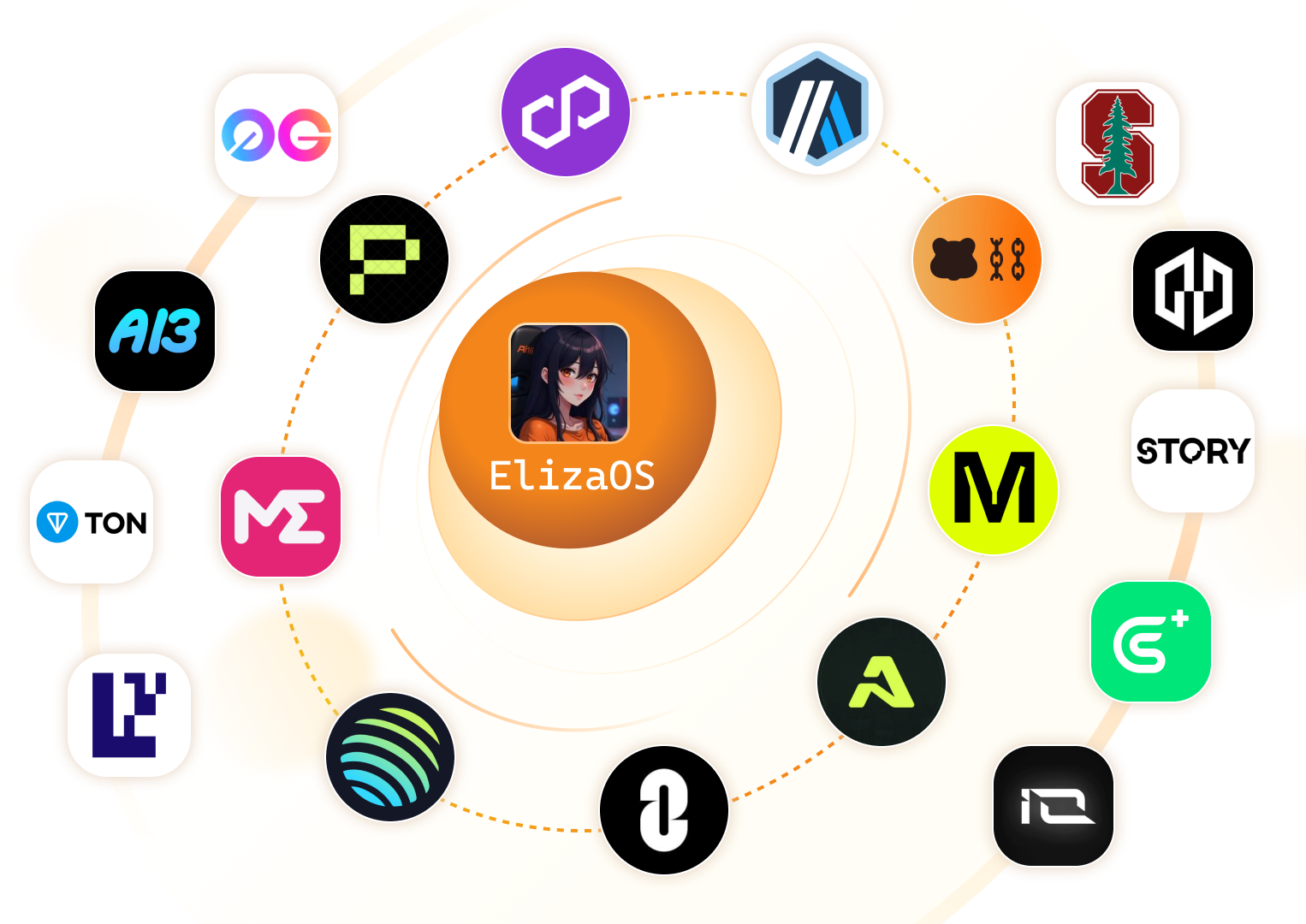}
  \caption{As of January 1, 2025, numerous representative web3 projects have built AI agents based on ElizaOS, with their combined market capitalization surpassing \textbf{\$20 billion dollars}.}
  \label{fig:dataset}
\end{figure}
Evaluating the success of ElizaOS's design philosophy and market penetration presents unique challenges in the rapidly evolving web3 AI agent ecosystem. To address this, we adopted a straightforward data-driven approach centered on strategic partnership analytics. Our methodology examined verified partnership announcements and collaborations since ElizaOS's July 2024 launch, comparing its adoption rate against leading web3 AI agent frameworks such as Myshell, Virtuals, and Swarm. \cref{fig:dataset} illustrates the ecosystem growth through a selection of web3 projects that leverage ElizaOS to construct their AI agent systems, with the cumulative market capitalization of ecosystem partners exceeding \textbf{\$20 billion dollars}. This substantial market presence underscores the growing confidence of developers and enterprises in ElizaOS's infrastructure.

\section{Limitation}
\label{sec: limitation}

Although Eliza offers significant advantages, it still has areas for improvement. The current limitations of Eliza can be categorized into three main areas: First, the absence of an \textbf{explicit workflow system} creates barriers for developers seeking to implement routine processes (\textit{e.g.}, periodic data summarization from multiple sources) within Eliza. For such requirements, GUI-enhanced workflow systems like Dify and Coze may be more suitable. Secondly, the \textbf{Runtime} design requires further refinement to balance the computational overhead of multiple agents—particularly as context and memory requirements scale exponentially—with runtime efficiency, especially in IO-intensive tasks. Thirdly, expanding multi-language support, particularly for languages such as Python and Rust, is essential for Eliza's future growth, as this would broaden its appeal to developers across different technological domains.

\section{Conclusion}

This paper introduces Eliza, a pioneering open-source Web3-friendly AI agent operating system designed to bridge the gap between AI technology and Web3 applications. By offering a platform that not only makes the deployment of Web3 applications possible but also effortless, Eliza stands out. We emphasize that every aspect of Eliza is crafted as a regular TypeScript program, ensuring it remains fully under user control while also providing seamless integration with Web3 functionalities, including but not limited to, reading and writing blockchain data and interacting with smart contracts.

The success of Eliza is attributed to its integration of strong Web3 demands into a design that balances utility and ease of use. We adhere to three main design principles: prioritizing Web3 developers, a pluggable modular design, and maintaining system simplicity while ensuring functionality. These principles have guided the development of Eliza, making it a powerful multi-agent system capable of interacting across multiple platforms.

The core concepts of ElizaOS, including Agents, Character Files, Providers, Actions, and Evaluators, together form a highly controllable and orchestrated framework focused on the Web3 industry, offering developers the tools to bring powerful AI agents to life. Our plugin architecture achieves modularity and extensibility while fostering community development and knowledge sharing.

Although Eliza offers significant advantages, we recognize areas for improvement. Current limitations include the absence of an explicit workflow system, the need for further refinement of the runtime design, and the necessity to expand multi-language support.

In summary, Eliza represents the forefront of a new era in technology, with possibilities limited only by the imagination of its users. As the fields of AI and Web3 continue to evolve rapidly, Eliza will continue to evolve to meet changing demands and pave the way for the future development of decentralized AI.

\section{Acknowlegments}

We express our gratitude to all individuals and organizations that have contributed to the development and refinement of Eliza. We thank all the Eliza core team members, contributors and package maintainers for their continuous effort and attic faith.


\clearpage
{\small
\bibliographystyle{plainnat}
\bibliography{main}
}

\clearpage

\end{document}